\documentclass[runningheads]{llncs}
\usepackage{graphicx}
\usepackage{comment}
\usepackage{amsmath,amssymb} % define this before the line numbering.
\usepackage{color}
\usepackage{caption} 
\usepackage{float}
\usepackage{hyperref}
\usepackage{subcaption} % for subfigures
\hypersetup{
    colorlinks=true,
    linkcolor=magenta
}
\pdfoutput=1
\begin{document}
% \linenumbers
\pagestyle{headings}
\mainmatter
\def\ECCVSubNumber{30}  % Insert your submission number here

\title{A Benchmark for Inpainting of Clothing Images with Irregular Holes} % Replace with your title

\begin{comment}
\titlerunning{ECCV-20 submission ID \ECCVSubNumber} 
\authorrunning{ECCV-20 submission ID \ECCVSubNumber} 
\author{Anonymous ECCV submission}
\institute{Paper ID \ECCVSubNumber}
\end{comment}
%******************

%\begin{comment}
\author{Furkan K{\i}nl{\i}\textsuperscript{1}
%\orcidID{0000-1111-2222-3333} 
\and
Bar{\i}\c{s} \"{O}zcan\textsuperscript{2}
%\orcidID{1111-2222-3333-4444} 
\and
Furkan K{\i}ra\c{c}\textsuperscript{3}
%\orcidID{2222--3333-4444-5555}
}
\authorrunning{K{\i}nl{\i} et al.}
\institute{\"{O}zye\u{g}in University, \.{I}stanbul, Turkey \\
\email{\{\textsuperscript{1}furkan.kinli, \textsuperscript{3}furkan.kirac\}@ozyegin.edu.tr}
\email{\textsuperscript{2}baris.ozcan.10097@ozu.edu.tr}}
%\end{comment}
%******************
\maketitle

\begin{figure}[!ht]
        \centering
        \begin{subfigure}[b]{0.24\textwidth}
                \includegraphics[width=0.9\textwidth]{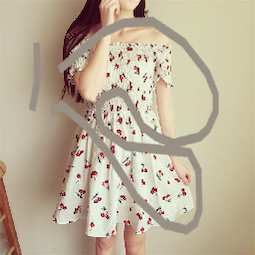}
                \includegraphics[width=0.9\textwidth]{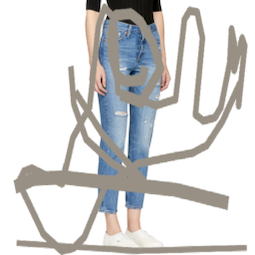}
                \caption{}
                \label{fig:firstpage0}
        \end{subfigure}       
        \begin{subfigure}[b]{0.24\textwidth}
                \includegraphics[width=0.9\textwidth]{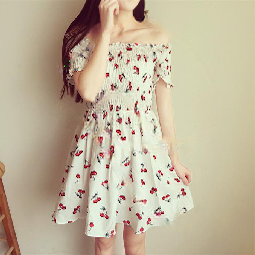}
                \includegraphics[width=0.9\textwidth]{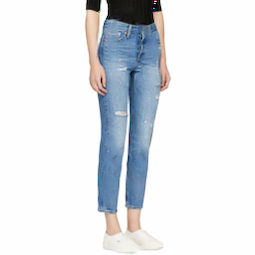}
                \caption{}
                \label{fig:firstpage1}
        \end{subfigure}
        \begin{subfigure}[b]{0.24\textwidth}
                \includegraphics[width=0.9\textwidth]{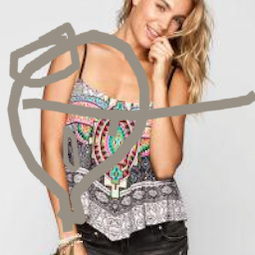}
                \includegraphics[width=0.9\textwidth]{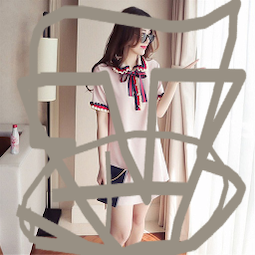}
                \caption{}
                \label{fig:firstpage2}
        \end{subfigure}
        \begin{subfigure}[b]{0.24\textwidth}
               \includegraphics[width=0.9\textwidth]{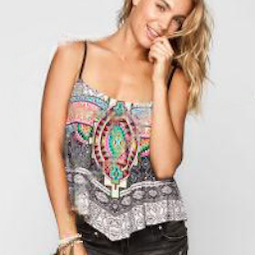}
               
               \includegraphics[width=0.9\textwidth]{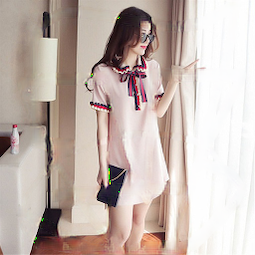}
                \caption{}
                \label{fig:firstpage3}
        \end{subfigure}
        
        \caption{Masked \& generated images of our model employing dilated partial convolutions for clothing image inpainting.}\label{fig:figure0} 
\end{figure}
\vspace{-1.1cm}
\begin{abstract}
Fashion image understanding is an active research field with a large number of practical applications for the industry. Despite its practical impacts on intelligent fashion analysis systems, clothing image inpainting has not been extensively examined yet. For that matter, we present an extensive benchmark of clothing image inpainting on well-known fashion datasets. Furthermore, we introduce the use of a dilated version of partial convolutions, which efficiently derive the mask update step, and empirically show that the proposed method reduces the required number of layers to form fully-transparent masks. Experiments show that dilated partial convolutions (DPConv) improve the quantitative inpainting performance when compared to the other inpainting strategies, especially it performs better when the mask size is 20\% or more of the image.
\end{abstract}

\section{Introduction}

Image inpainting is the task of filling the holes in a particular image with some missing regions in such a way that the generated image should be visually plausible and semantically coherent. There are numerous vision applications including object removal, regional editing, super-resolution, stitching and many others that can be practicable by employing different image inpainting strategies. In the literature, the most prominent studies \cite{patchmatch,contextencoders,localandglobal,contextual,partial,gated,edgeconnect} proposing different strategies for solving inpainting problems have focused on mostly natural scene understanding, street view understanding and face completion tasks. 

Fashion image understanding is an active research field that has enormous potential of practical applications in the industry. With the achievements of deep learning-based solutions \cite{resnet,rcnn,maskrcnn,unet} and increasing the number of datasets representing more real-world-like cases \cite{fashiongen,fashionai,deepfashion,deepfashion2}, different solutions have been proposed for various image-related vision tasks in fashion domain such as clothing category classification \cite{deepfashion,deepfashion2,afgn,fmula,fashioncapsnet}, attribute recognition \cite{deepfashion,deepfashion2,capsulewardrobe,mixandmatch,taam}, clothing segmentation \cite{coparsing,deepfashion2,ssfpn,sladn,gmss}, clothing image retrieval \cite{vrsfi,stylefinder,darn,vam,bier,rccapsnet} and clothing generation \cite{design,capsulewardrobe,viton,uvviton,laviton,zhu,gunel,yildirim,attrgan,han}. At this point, the recent breakthroughs in deep generative learning solutions \cite{gan,stylegan,progan} lead to arise some opportunities for the applications in fashion domain including designing new clothing items for recommendation systems \cite{design,capsulewardrobe}, virtual try-on systems \cite{viton,uvviton,laviton} and fashion synthesis \cite{zhu,gunel,attrgan,yildirim,han}. Despite these extensive studies on different tasks in fashion domain, image inpainting has not been extensively examined yet. Considering the practical impacts of achieving this task on real-world applications of fashion analysis, we present an extensive benchmark for clothing image inpainting by employing a generative approach that recovers the irregular missing regions in the set of clothing items. 

\begin{figure}[!ht]
        \centering
        \begin{subfigure}[b]{\textwidth}
                \includegraphics[width=0.995\textwidth]{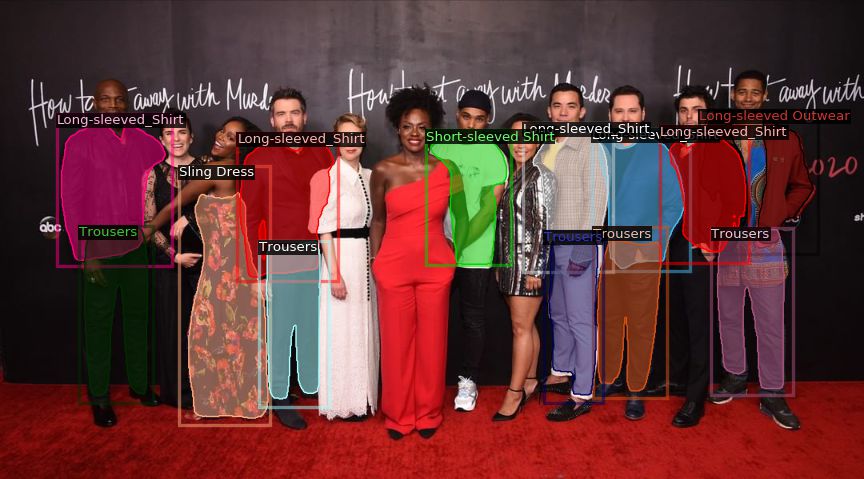}
                \includegraphics[width=0.3125\textwidth]{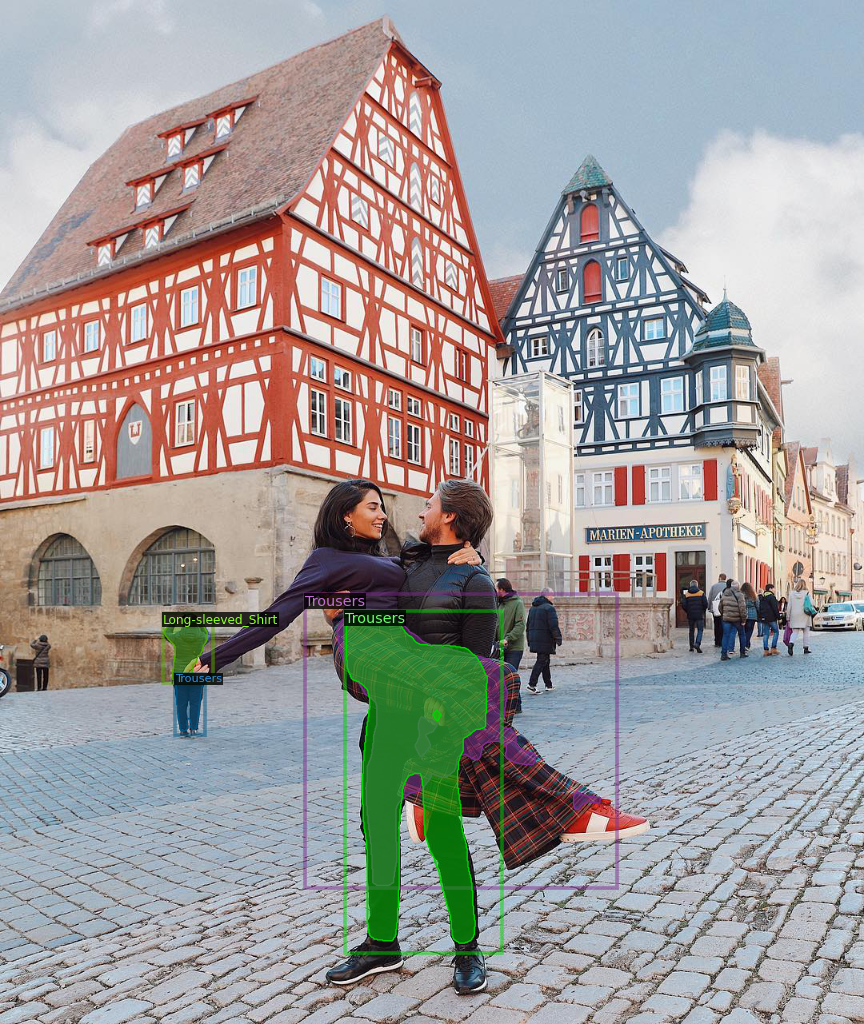}
                \includegraphics[width=0.375\textwidth]{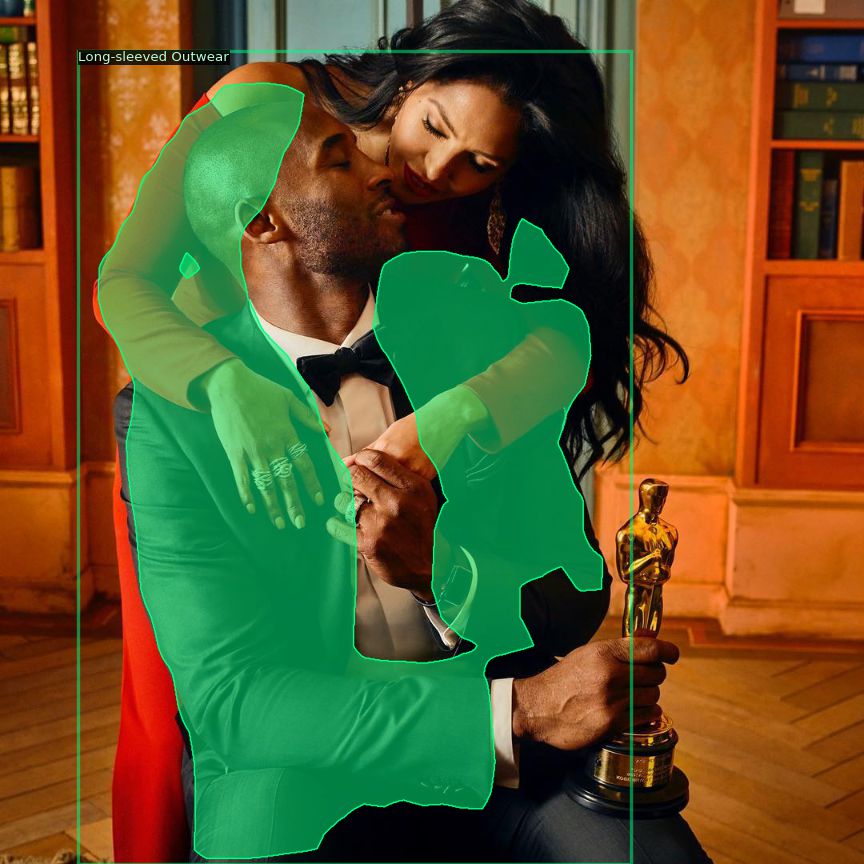}
                \includegraphics[width=0.3\textwidth]{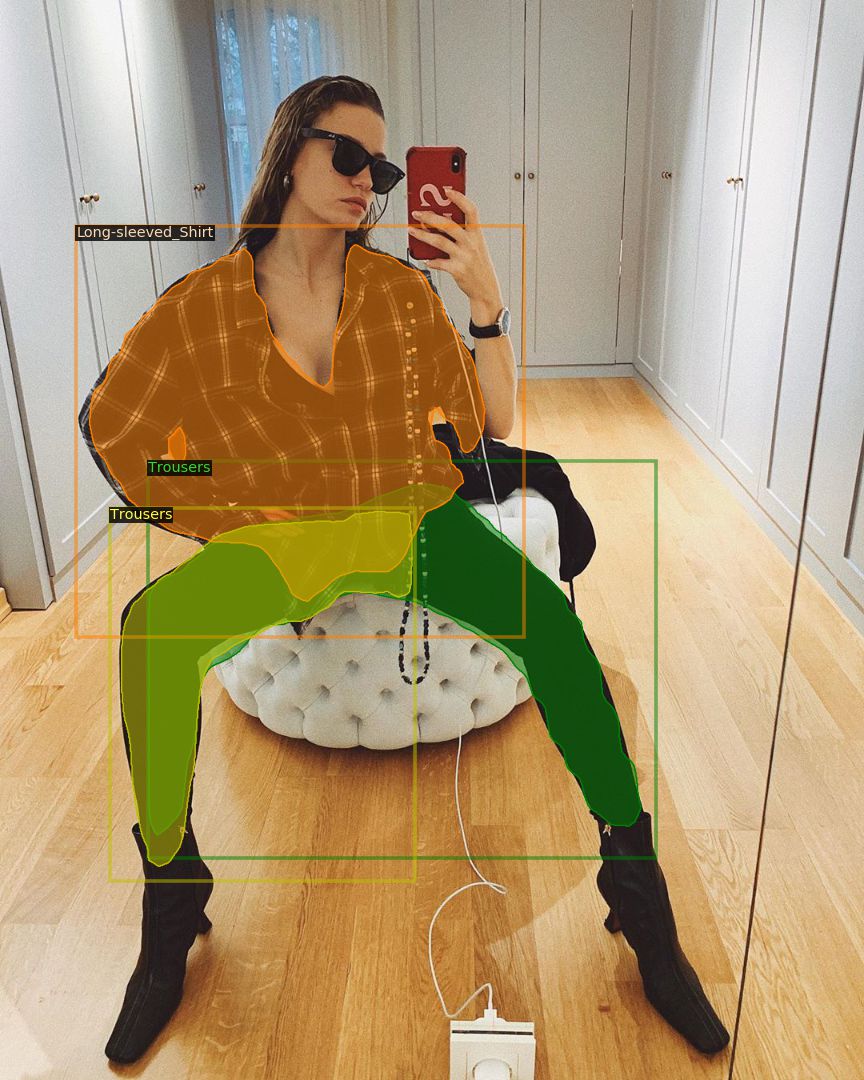}
                \caption{}
        \end{subfigure}       
        
        \caption{Difficult scenarios for fashion image understanding due to the occlusions (natural, multiple persons, multiple clothing items)}\label{fig:figure1} 
\end{figure}

To extend the practical advantages of this solution for fashion domain, we need further investigation of the drawbacks of the current fashion analysis systems working on real-world cases. Due to possible deformations and occlusions in real-world scenarios, understanding fashion images is a challenging task to make successful inferences. For instance, first, attribute recognition models trained with commercial images often fail to successfully infer the attributes of social media images (in other words, consumer images), due to the occlusions that appears on the parts representing the actual attribute of a particular clothing. Next, overlapping items in a single clothing image builds natural occlusion scenarios for both items when segmentation is applied to the image. Although such segmentation models \cite{detectron,detectron2,mmdetection} have the ability to infer the possible locations of the occluded region of an object, filling this region in a visually plausible and semantically correct way in order to analyze the segmented clothing items is yet to be achieved. Moreover, for a visual recommendation system working with real-world clothing images, any deformation on the extracted clothing items can change the possible combinations of it with the other items. Based on such drawbacks on fashion analysis systems, applying different inpainting strategies may be practical for understanding fashion images better. Therefore, we want to contribute to the fundamental research effort of solving inpainting problems by redirecting it to fashion domain.

In image inpainting literature, there are several strategies that use image statistics or deep learning-based solutions. PatchMatch \cite{patchmatch} is one of the most prominent strategies relying on the image statistics, and it basically searches for the best fitting patch to fill the rectangular-formed missing parts in the images. Although it is possible to generate visually plausible results by using this inpainting strategy, it cannot achieve semantic coherence in hard scenarios since it only depends on the statistics of available parts in the image. On the other hand, deep learning-based solutions are more suitable for image inpainting since deep neural networks can inherently learn the hidden representations by preserving semantic priors. Earlier studies \cite{contextencoders,localandglobal,contextual} try to solve the problem of initialization of the pixels in the missing parts to condition the output by assigning a fixed value for these pixels. In these studies, the results have the visual artifacts, and need some additional post-processing methods (\textit{e.g.} fast marching \cite{fastmarching}, Poisson image blending \cite{poisson} and the refinement network \cite{contextual}) to refine them. Partial convolutional mechanism \cite{partial} addresses this limitation by conditioning the only valid pixels. To achieve this, convolutional layers are masked and re-weighted, and then the mask is updated in such a way that progressively filling up the hole pixels. Recently, Nazeri \textit{et al.} \cite{edgeconnect} considers inpainting problem as image completion by predicting the structure (\textit{i.e.} edge maps) of the main content in the image. Apart from these studies, we focus on increasing the receptive field size of low-resolution feature maps of partial convolutions, and thus, we propose dilated partial convolutions whose the kernel of partial convolution \cite{partial} is dilated by a given parameter, and capable of gathering more information from near-to-hole regions in order to update the masks more efficiently without requiring to decrease their spatial dimensionality up to very low-resolution.

In summary, our main contributions are as follows:
\begin{itemize}
    \item We introduce an extensive benchmark of clothing image inpainting on a variety of challenging datasets including FashionGen \cite{fashiongen}, FashionAI \cite{fashionai}, DeepFashion \cite{deepfashion} and DeepFashion2 \cite{deepfashion2}, and attempt to redirect the fundamental research efforts on image inpainting problems to fashion domain.
    \item To the best of our knowledge, this is the first study to mention the practical usefulness of achieving visually plausible and semantically coherent inpainting of fashion images for industrial applications.
    \item We present the idea of using dilated version of partial convolutions for image inpainting tasks, where the dilated input window for partial convolutions derives more efficient mask update step.
    \item We empirically show that dilated partial convolutions reduce the number of layers that requires to lead to a mask without any holes, and thus, it makes possible to achieve better inpainting quality without reducing the spatial dimensionality of encoder output. 
\end{itemize}{}

\section{Related Work}

Image inpainting is a challenging task, and it has been extensively studied for a decade in different vision-related research fields, especially on natural scene understanding and face recognition. Several traditional inpainting strategies such as \cite{ballester,bertalmio,efros,hays,fastmarching} try to synthesize texture information in the holes by employing available image statistics. However, these methods often fail to achieve inpainting the images without any artifacts since using only available image statistics may mislead the results about the texture information between holes and non-hole regions when it varies. Barnes \textit{et al.} \cite{patchmatch} proposes a fast algorithm, namely \textit{PatchMatch}, which iteratively searches for the best fitting patches to the missing parts of the image. Although this method can produce visually plausible results in a faster way, it still lacks of producing semantically coherent results, and far from being suitable for real-time image processing pipeline.

In deep learning-based solutions, earlier approaches use a constant value for initializing the holes before passing throughout the network. First, \textit{Context Encoders} \cite{contextencoders} employs an Auto-encoder architecture with adversarial training strategy in order to fill a central large hole in the images. Yang \textit{et al.} \cite{multiscale} extends the idea in \cite{contextencoders} with post-processing the output that considers only the available image statistics. Song \textit{et al.} \cite{coarsetofine} proposes a robust training scheme in coarse-to-fine manner. Iizuka \textit{et al.} \cite{localandglobal} introduces a different adversarial training strategy to provide both local and global consistency in the generated images. Also, in \cite{localandglobal}, it is demonstrated that sufficiently larger receptive fields work well in image inpainting tasks, and dilated convolution is adopted for increasing the size of the receptive fields. Yu \textit{et al.} \cite{contextual} improves the idea of using dilated convolutions in inpainting task by adding contextual attention mechanism on top of low-resolution feature maps for explicitly matching and attending to relevant background patches. Liu \textit{et al.} \cite{partial} presents partial convolutions where the convolution is masked and re-weighted to be conditioned on only valid pixels. Yu \textit{et al.} \cite{gated} proposes gated convolutions that have the ability to learn the features dynamically at each spatial location across all layers, and this mechanism improves the color consistency and semantic coherence in the generated images. Liu \textit{et al.} \cite{csa} employs coherent semantic attention and consistency loss to the refinement network to construct the correlation between the features of hole regions, even in deeper layers. K{\i}nl{\i} \textit{et al.} \cite{kinli2020inpainting} observes the effect of collaborating with textual features extracted by image descriptions on inpainting performance. Nazeri \textit{et al.} \cite{edgeconnect} proposes a novel method that predicts the image structure of the missing region in the form of edge maps, and these predicted edge maps are passed to the second stage to guide the inpainting.

\section{Methodology}

Our proposed model enhances the capability of partial convolutions \cite{partial}, which alters this mechanism by adding a dilation factor to its convolutional filters, and employing self-attention to the decoder part of our model. In this section, we first explain the reasoning behind dilated partial convolutions and present its formulation. Then, we describe our architecture, and lastly discuss the loss functions of proposed model.

\subsection{Dilated Partial Convolutions}
Unlike in natural images where spatially-near pixels yield a larger correlation, for clothing image inpainting, the correlated pixels may be far apart in a particular image (\textit{e.g.} a pattern on a shirt can cover a larger area of the image, and more importantly, this pattern may not be spatially continuous in such scenarios with occlusions or deformations). Even though both partial and dilated convolutional layers prove that they can produce visually plausible results in inpainting tasks, they have their own particular shortcomings. When partial convolutions are employed, the layers cannot gather the information from the correlation of far apart pixels as in dilated convolutions, on the other hand, the network does not only focus on non-hole regions of the images by using only dilated convolutions \cite{dilated}. We address both these shortcomings by introducing the use of dilated partial convolutions where the input window is dilated by a given parameter, and the mask of partial convolution is applied afterwards. This allows the network to focus on non-hole regions, and it has larger receptive fields to utilize correlations of far apart pixels. More importantly, as stated in \cite{partial}, the consecutive layers of partial convolutions will eventually lead to a mask without any holes depending on the input mask, and we have empirically proven that dilated partial convolutions reduce the number of required layers to achieve this, and the input mask covers up a larger percentage of the input image in earlier part of the network. The discussion and detailed empirical results can be found in Discussion part. The formal definition of dilated partial convolutions can be formulated as follows:

\begin{equation}
    \label{eq:M}
    \mathcal{M} = \sum\limits_{m=-M}^M \sum\limits_{n=-N}^N m(x-ln,y-lm)
\end{equation}

\begin{equation}
    \label{eq:nfactor}
    Z = \frac{(2M+1) \times (2N+1)}{\mathcal{M}} 
\end{equation}

\begin{equation}
    \label{eq:pdconv}
    (f_m \circledast g_l)(x,y) = \frac{\sum\limits_{m=-M}^M \sum\limits_{n=-N}^N f(x-ln,y-lm) g(n,m) m(x-ln,y-lm)} {Z}
\end{equation}
where $f$, $g$ and $m$ represents the input, the convolutional kernel with a size of $(2M+1) \times (2N+1)$ and the corresponding mask, respectively. Eq. \eqref{eq:nfactor} has essentially the same scaling factor as in \cite{partial}, but modified to be applicable to dilated partial convolutions. Note that when the dilation factor $l=1$, Eq. \eqref{eq:pdconv} becomes the partial convolution without any dilation. Mask update is calculated in the same manner as in partial convolutions, and can be formulated as follows:

\begin{equation}
    m' =   \begin{cases}
            1,  &\text{if \(\mathcal{M} > 1\)} \\
            0,  &\text{otherwise}
            \end{cases}
\end{equation}
where the updated mask $m'$ gets closer to become fully-transparent (\textit{i.e.} all pixels become ones) after a certain number of layers. Dilated partial convolutions allows $m'$ to become fully-transparent in earlier layers when compared to partial convolutions.

\begin{figure}[!t]
\includegraphics[width=\textwidth]{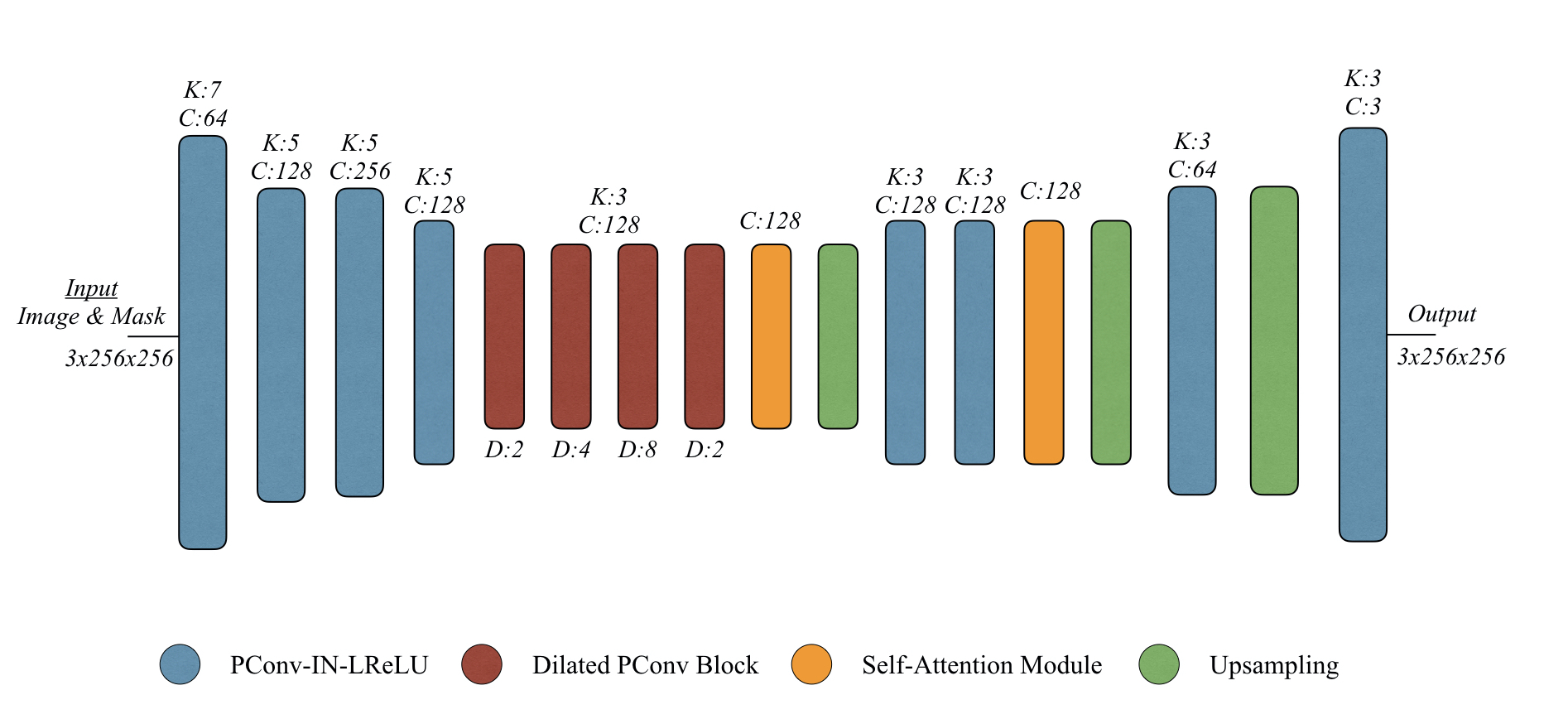}

\caption{Overview of our architecture design.}
\label{fig:figure2}
\end{figure}

\subsection{Model Architecture}
We have designed a similar \textit{U-Net-like} \cite{unet} architecture as in \cite{partial}, where all low resolution layers are replaced with dilated partial convolutions while the first four layers are left as partial convolutions. The binary mask in the first layer is defined as the input corruption mask. Moreover, in the decoder stage, we have used self-attention module \cite{selfattention} to make use of spatially distant but related features. All of the skip connections are concatenated with the layer on corresponding level, instead of adding to them, and also dilated partial convolutions are residually-connected. % and decoder stage use regular convolutions. 
The dilation rate is progressively increased up to the last dilated convolutional layer (\textit{i.e.} multi-scale context aggregation), as in \cite{dilated}. As working on the fashion datasets, the input sizes for all models are picked as $256 \times 256$. Overall architecture design can be seen in Figure \ref{fig:figure2}. 

\subsection{Loss Functions}
We have followed a similar loss function scheme to \cite{partial}, but with slight differences. The total loss function is introduced in Eq. \eqref{eq:TotalLoss}, where $\mathcal{L}_{pixel}$ is the pixel loss, $\mathcal{L}_{style}$ is the style loss, $\mathcal{L}_{adv}$ is the adversarial loss and $\mathcal{L}_{tv}$ is the total variation (TV) loss. We found that the coefficients in Eq. \eqref{eq:TotalLoss} work better than the ones in \cite{partial} in our experimental settings. Moreover, the perceptual loss is substituted with an adversarial loss since we experimentally found that the benefits of the perceptual loss do not afford its computational cost, and also we want to take the advantage of adversarial training for our model.

\begin{equation}
    \label{eq:TotalLoss}
    \mathcal{L}_{total} = 10 \mathcal{L}_{pixel} + 120 \mathcal{L}_{style} + 10^{-3} \mathcal{L}_{adv} + 10^{-4}\mathcal{L}_{tv}
\end{equation}

Pixel loss $\mathcal{L}_{pixel}$ is the sum of $\ell_{1}$ losses of the hole and non-hole regions between the output image $\mathbf{I_\textit{out}}$ and the ground truth image $\mathbf{I}_{gt}$, as given in \eqref{eq:pixelL}, where $ N_{\mathbf{I}_{gt}} = C \times H \times W$ as $C, H$ and $W$ are the input dimensions, and $\hat{m}$ denotes the initial mask on the input.

\begin{equation}
    \label{eq:pixelL}
    \mathcal{L}_{pixel} = \frac{1}{N_{\mathbf{I}_{gt}}} (\bigg\| (1 - \hat{m})\odot (\mathbf{I}_{out} - \mathbf{I}_{gt}) \bigg\|_{1}  + \bigg\| \hat{m}\odot (\mathbf{I}_{out} - \mathbf{I}_{gt}) \bigg\|_{1})
\end{equation}{}

The total style loss $\mathcal{L}_{style}$ is calculated by summing the style losses for $\mathbf{I}_{out}$ and $\mathbf{I}_{comp}$, denoted as $\mathcal{L}_{style_{out}}$ and $\mathcal{L}_{style_{comp}}$, respectively (Eq. \eqref{eq:styleL}). To obtain the composite image $\mathbf{I}_{comp}$, all non-hole pixels in the output image are replaced with the ground truth pixels. Style loss requires to calculate the activation maps ${ \Psi^{ \mathbf{I}_{\ast} } }_{p}$ of pre-trained VGG-16 with respect to the input image $\mathbf{I}_{\ast}$, where $p$ denotes the layer index. For a given layer index $p$, the normalization factor is defined as $K_p = \frac{1}{C_pH_pW_p}$.

%\begin{equation}
%    \label{eq:ostyleL}
%    \mathcal{L}_{style_{out}}  = \sum\limits_{p=0}^{P-1} \frac{1}{C_p^2} \bigg\| K_p(({\Psi^{\mathbf{I}_{out}}}_{p})^T ({\Psi^{\mathbf{I}_{out}}}_{p}) - ({\Psi^{\mathbf{I}_{gt}}}_{p})^T ({\Psi^{\mathbf{I}_{gt}}}_{p}) )   \bigg\|_1
%\end{equation}

%\begin{equation}
%    \label{eq:cstyleL}
%    \mathcal{L}_{style_{comp}}  = \sum\limits_{p=0}^{P-1} \frac{1}{C_p^2} \bigg\| K_p(({\Psi^{\mathbf{I}_{comp}}}_{p})^T ({\Psi^{\mathbf{I}_{comp}}}_{p}) - ({\Psi^{\mathbf{I}_{gt}}}_{p})^T ({\Psi^{\mathbf{I}_{gt}}}_{p}) )   \bigg\|_1
%\end{equation}

\begin{equation}
    \label{eq:styleL}
    \begin{aligned}
    \mathcal{L}_{style} ={} &\sum\limits_{p=0}^{P-1} \frac{1}{C_p^2} \bigg\| K_p(({\Psi^{\mathbf{I}_{out}}}_{p})^T ({\Psi^{\mathbf{I}_{out}}}_{p}) - ({\Psi^{\mathbf{I}_{gt}}}_{p})^T ({\Psi^{\mathbf{I}_{gt}}}_{p}) )   \bigg\|_1 + \\ &\sum\limits_{p=0}^{P-1} \frac{1}{C_p^2} \bigg\| K_p(({\Psi^{\mathbf{I}_{comp}}}_{p})^T ({\Psi^{\mathbf{I}_{comp}}}_{p}) - ({\Psi^{\mathbf{I}_{gt}}}_{p})^T ({\Psi^{\mathbf{I}_{gt}}}_{p}) )   \bigg\|_1
    \end{aligned}
\end{equation}{}

The adversarial loss given in Eq. \eqref{eq:advLG} is adopted from \cite{gan}, where $\mathcal{G}$ is the generator network, $\mathcal{D}$ is the discriminator network, and $\mathbf{I}^M$ is the masked input image. The discriminator $\mathcal{D}$ network is a simple CNN with a depth of 5, and trained to optimize  the loss function given in Eq. \eqref{eq:advLD}. To train the discriminator better, we flipped the labels with the probability of 0.1 to make the labels noisy for the discriminator, and also we applied label smoothing where, for each instance, if it is real, then replace its label with a random number between 0.7 and 1.2, otherwise, replace it with 0.0 and 0.3. Lastly, we tried to apply random dropout for the decoder part of our model.

\begin{equation}
    \label{eq:advLG}
    \mathcal{L}_{a d v_{G}}:=\mathbb{E}\left[\left(\mathcal{D}\left(\mathcal{G}\left(\mathbf{I}^{M}\right) -1\right)\right)^{2}\right]
\end{equation}

\begin{equation}
    \label{eq:advLD}
    \mathcal{L}_{a d v_{D}}:=\mathbb{E}\left[\mathcal{D}(\hat{\mathbf{I}})^{2}\right]+\mathbb{E}\left[\left(\mathcal{D}\left(\mathbf{I} \right)-1\right)^{2}\right]
\end{equation}{}

Lastly, total variation loss $L_{tv}$, as given in \eqref{eq:tvL}, enforces the spatial continuity on the generated images, where $R$ is the region after applying a 1-pixel dilation to a hole region in the input image.

\begin{equation}
    \label{eq:tvL}
    \mathcal{L}_{t v}=\sum_{\scriptscriptstyle(i, j) \in R,(i, j+1) \in R} \frac{\left\|\mathbf{I}_{c o m p}^{i, j+1}-\mathbf{I}_{c o m p}^{i, j}\right\|_{1}}{N_{\mathbf{I}_{c o m p}}}+\sum_{\scriptscriptstyle(i, j) \in R,(i+1, j) \in R} \frac{\left\|\mathbf{I}_{c o m p}^{i+1, j}-\mathbf{I}_{c o m p}^{i, j}\right\|_{1}}{N_{\mathbf{I}_{c o m p}}}
\end{equation}

\section{Experiments}

In this study, we introduce an extensive benchmark on fashion image inpainting, and also propose enhanced version of partial convolutions, namely \textit{dilated partial convolutions}. We have conducted our experiments on four different well-known fashion datasets, which are FashionGen \cite{fashiongen}, FashionAI \cite{fashionai}, DeepFashion \cite{deepfashion} and DeepFashion2 \cite{deepfashion2}.

\subsection{Experimental Setup}

\subsubsection{Training details} 

In our experiments, we used Adam optimizer \cite{adam} with $\beta_{1} = 0.5$ and $\beta_{2} = 0.9$ for both generator and discriminator networks. The initial learning rate for generator network is $2 \times 10^{-4}$, and for discriminator network, is $10^{-4}$. We trained our model for {\raise.17ex\hbox{$\scriptstyle\sim$}}120.000 steps for each dataset with batch size of 64. We only applied horizontal flipping to the input images, no other data augmentation technique is applied. For DeepFashion and DeepFashion2 datasets, the size of images varies, so we use random cropping (central cropping for testing) and resizing to obtain the size of $256 \times 256$ for training images. We implemented our framework with PyTorch library \cite{pytorch}, and used 2x NVIDIA Tesla V100 GPUs for our training. Source code: \url{https://github.com/birdortyedi/fashion-image-inpainting}

\subsubsection{Datasets}

To create an extensive benchmark for fashion image inpainting, we picked four common fashion datasets to use in our experiments. Before training, we prepared the datasets to the image inpainting setup by applying inpainting masks\footnote{\url{https://github.com/karfly/qd-imd}} to the images, which erase some parts randomly. To make sure that the images have some holes on clothing parts, we generated the masks with a heuristic, where at least a small portion of the clothing parts has been erased by the mask. Then, we run our experiments for each method on each dataset separately.

\textbf{FashionGen (FG)} \cite{fashiongen} contains several different fashion products collected from an online platform selling the luxury goods from independent designers. Each product is represented by an image, the description, its attributes, and relational information defined by professional designers. We extract the instances belonging to these categories from the dataset. At this point, training set has 200K clothing images from 9 different categories (\textit{top, sweater, pant, jean, shirt, dress, short, skirt, coat}), and validation set has 30K clothing images.

\textbf{FashionAI (FAI)} \cite{fashionai} is introduced in FashionAI Global Challenge 2018 by The Vision \& Beauty Team of Alibaba Group and Institute of Textile \& Clothing of The Hong Kong Polytechnic University. For this dataset, there are 180K consumer images of different clothing categories for training, and 10K for testing.

\textbf{DeepFashion (DF)} \cite{deepfashion} is a dataset containing {\raise.17ex\hbox{$\scriptstyle\sim$}}800K diverse fashion images with their rich annotations (\textit{i.e.} 46 categories, 1.000 descriptive attributes, bounding boxes and landmark information) ranging from well-posed product images to real-world-like consumer photos. In our experiments on this dataset, we follow the same procedure as in \cite{deepfashion} to split training and testing sets, so we have 207K clothing images for training, and 40K for testing.

\textbf{DeepFashion2 (DF2)} \cite{deepfashion2} is one of the largest open-source fashion database that contains 491K high-resolution clothing images with 13 different categories and numerous attributes. Similar to DeepFashion, we again follow the procedure in the paper of dataset \cite{deepfashion2} to split the published version of the data.

\subsection{Quantitative Results}

\begin{table}[!t]
\caption{Quantitative comparison between our proposed model and the state-of-the-art methods on four well-known fashion datasets.}
\resizebox{\columnwidth}{!}{
\centering
\begin{tabular}{l|l|l|l|l|l|l|l|l|l|l|l|l}
%\toprule
                \textbf{Mask Ratios} & \multicolumn{4}{c|}{\textbf{[0.0:0.2]}}                                                                       & \multicolumn{4}{c|}{\textbf{[0.2:0.4]}}                                                                        & \multicolumn{4}{c}{\textbf{[0.4:]}}                                                                \\ \hline \hline
               \textbf{Datasets} & \multicolumn{1}{c|}{FG} & \multicolumn{1}{c|}{FAI} & \multicolumn{1}{c|}{DF} & \multicolumn{1}{c|}{DF2} & \multicolumn{1}{c|}{FG} & \multicolumn{1}{c|}{FAI} & \multicolumn{1}{c|}{DF} & \multicolumn{1}{c|}{DF2} & \multicolumn{1}{c|}{FG} & \multicolumn{1}{c|}{FAI} & \multicolumn{1}{c|}{DF} & \multicolumn{1}{c}{DF2} \\ 
%\midrule 
\hline \hline 
$\ell_{1}$ (PM) \textit{\%}      &            \textbf{0.66}            &            1.16              &      1.34                  &      1.32                    &     1.34                   &            2.11             &      2.03                  &    1.96                    &   3.70                     &       6.12                 &      5.74                 &         5.43               \\
$\ell_{1}$ (PConv) \textit{\%}   &         0.73               &       0.92                  &       0.91                &        0.94                &              1.18           &         1.48                &              1.34          &           1.47               &         2.82               &           4.86              &      3.76                &          3.70          \\
$\ell_{1}$ (GConv) \textit{\%}   &      0.76              &       0.95                  &           0.93            &      0.95                   &       1.22                 &      1.58                   &       1.23               &            1.36              &       2.99                 &      5.20                  &         3.65          &             3.62           \\
$\ell_{1}$ (\textbf{DPConv}) \textit{\%}  &        \t
0.70                &         \textbf{0.91}                &        \textbf{0.87}                &         \textbf{0.92}                  &            \textbf{1.14}            &           \textbf{1.39}              &   \textbf{1.20}                    &          \textbf{1.33}                &         \textbf{2.61}                &        \textbf{4.03}                 &       \textbf{3.36}                &      \textbf{3.38}                   \\ 
% \midrule 
\hline \hline
PSNR (PM)       &     16.14                   &         12.21                &          12.30              &                   12.55       &       14.71                 &           11.23              &        11.37                &             11.49             &       13.36                 &       9.74                  &       9.96                 &       10.09                  \\
PSNR (PConv)    &          16.04              &          \textbf{12.33}               &              12.43          &           12.73               &        15.04                &       11.49                  &       11.66                &            11.94              &           13.91             &                 11.11        &             11.47           &    11.89                     \\
PSNR (GConv)    &           15.98             &         12.20                &         12.34               &             12.66             &            14.86            &        11.42                 &     11.61                   &             11.84             &      13.79                  &      11.02                   &     11.39                  &           11.77             \\
PSNR (\textbf{DPConv})   &          \textbf{16.23}              &                   12.27      &          \textbf{12.56}              &       \textbf{12.89 }              &  \textbf{15.15}                       &            \textbf{11.81}          &            \textbf{11.88}             &      \textbf{12.12}                    &                 \textbf{14.16}       &       \textbf{11.63}                  &      \textbf{11.84}                  &    \textbf{12.28}                     \\ 
%\midrule 
\hline \hline
SSIM (PM)       &           \textbf{0.891}             &             0.773          &                  0.784      &       0.804                 &       0.816                 &           0.737              &       0.744                 &           0.769               &              0.738          &          0.622               &                0.671        &        0.677                 \\
SSIM (PConv)    &            0.851            &          \textbf{0.784}              &            0.777            &          0.804                &       0.770                  &     0.735                    &      0.729                  &          0.733                &                0.721        &              0.649           &           0.689             &        0.688                 \\
SSIM (GConv)    &           0.845             &          0.751               &          0.768            &            0.796              &              0.783          &         0.723                &      0.731                 &          0.741                &    0.717                    &     0.624                   &           0.670              &        0.681                 \\
SSIM (\textbf{DPConv})   &         0.855        &                0.782         &            \textbf{0.794}            &           \textbf{0.811}              &     \textbf{0.824}               &               \textbf{0.739}          &       \textbf{0.750}                 &        \textbf{0.777}                  &    \textbf{0.766}                    &       \textbf{0.683}                  &      \textbf{0.717}                  &        \textbf{0.719}                 \\ 
%\midrule 
\hline \hline
FID (PM)     &           \textbf{2.99}             &           5.44              &           \textbf{4.36}              &           5.78                &            6.21             &                  11.98        &        11.65               &          12.30                &         25.92               &         28.91                 &         27.94               &          29.74               \\
FID (PConv)  &         3.76              &           6.04               &          5.93              &        6.22                 &          5.56              &           11.05              &                     10.36   &           11.28               &          23.47              &         27.42               &          24.39              &          27.52               \\
FID (GConv)  &         3.57               &         5.86                &          5.63              &         6.18                  &         5.28               &           10.84              &                   10.12     &           11.01                &         23.18               &        27.19                 &      24.21                  &            26.95             \\
FID (\textbf{DPConv}) &        3.18                &          \textbf{5.24}              &         4.66               &         \textbf{5.48}            &             \textbf{5.20}           &          \textbf{10.08}              &                \textbf{9.26}        &          \textbf{10.21}               &          \textbf{22.82}              &          \textbf{25.55}               &         \textbf{23.94}                &         \textbf{25.62}                \\ 
%\bottomrule 
\hline \hline
\end{tabular}
}
\label{tab:quantitative}
\end{table}

As is the case with the previous studies \cite{contextual,partial,gated,csa}, we evaluate the performance of our model with four different metrics, which are $\ell_{1}$ error percentage, SSIM \cite{ssim}, PSNR and FID \cite{fid}. Then, we compare the quantitative performance of DPConv with PatchMatch (PM) \cite{patchmatch} as a non-learning method, Partial Convolutions (PConv) \cite{partial} and Gated Convolutions (GConv) \cite{gated}, as learning-based methods. We used the third-party implementations for PM\footnote{\url{https://github.com/MingtaoGuo/PatchMatch}} and GConv\footnote{\url{https://github.com/avalonstrel/GatedConvolution\_pytorch}}, but re-implemented PConv according to the layer implementation\footnote{\url{https://github.com/NVIDIA/partialconv}} and the architecture details referred in their paper \cite{partial}. The quantitative results of both our model and the state-of-the-art methods/models are shown in Table \ref{tab:quantitative}. The observations are as follows: (1) Dilated partial convolutions are more robust to the changes in the mask ratios. (2) According to SSIM metric, although PM performs well in the cases where the mask ratio is smaller, the performances on all measurements significantly decrease for more complex cases. (3) The performances of all methods are very similar on less complex dataset (\textit{i.e.} FashionGen), whereas the impact of our proposed model can be clearly observed as mask ratio increases in the settings of more complex datasets (\textit{i.e.} DeepFashion and DeepFashion2). (4) Using dilated version of partial convolutions in image inpainting improves the overall performance without reducing the spatial dimensionality of the encoder output. (5) Overall, our model, namely \textit{DPConv}, outperforms the current inpainting strategies on four well-known fashion datasets.

\subsection{Qualitative Results}

\captionsetup[subfigure]{labelfont=bf, labelformat=parens}
\begin{figure}[!t]
        \centering
        \begin{subfigure}[b]{0.15\textwidth}
                \includegraphics[width=\textwidth]{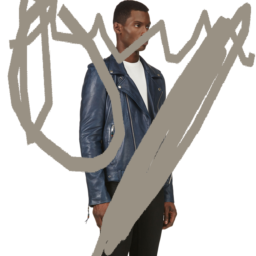}
                \includegraphics[width=\textwidth]{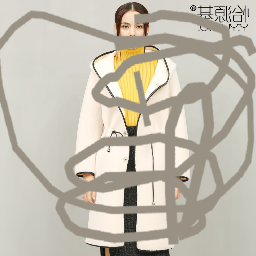}
                \includegraphics[width=\textwidth]{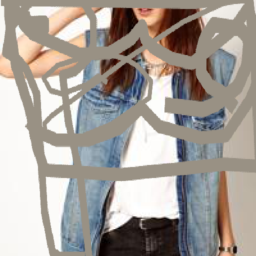}
                \includegraphics[width=\textwidth]{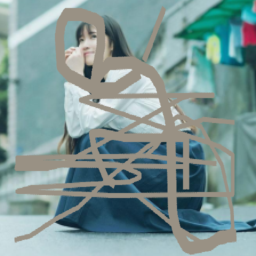}
                \caption{IN}
                \label{fig:lab0}
        \end{subfigure}       
        \begin{subfigure}[b]{0.15\textwidth}
                \includegraphics[width=\textwidth]{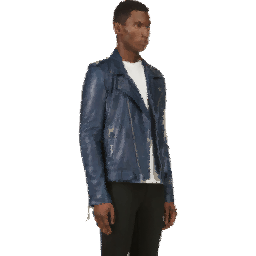}
                \includegraphics[width=\textwidth]{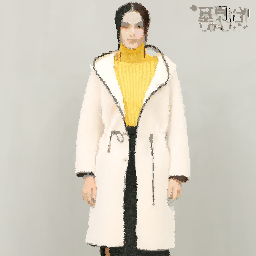}
                \includegraphics[width=\textwidth]{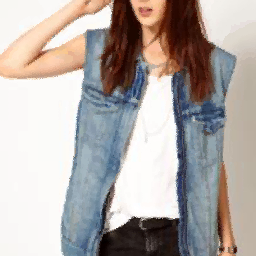}
                \includegraphics[width=\textwidth]{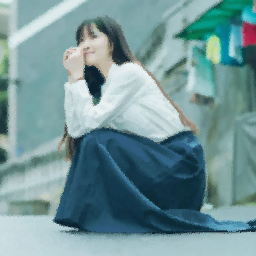}
                \caption{PM}
                \label{fig:lab1}
        \end{subfigure}
        \begin{subfigure}[b]{0.15\textwidth}
                \includegraphics[width=\textwidth]{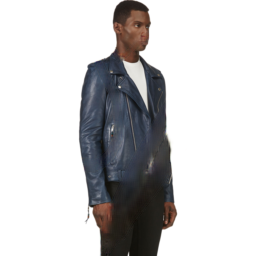}
                \includegraphics[width=\textwidth]{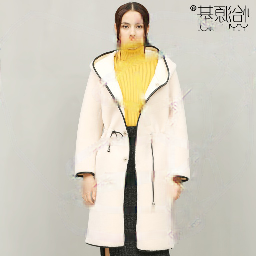}
                \includegraphics[width=\textwidth]{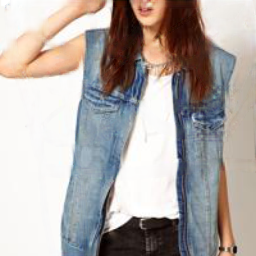}
                \includegraphics[width=\textwidth]{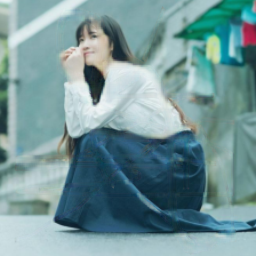}
                \caption{Partial}
                \label{fig:lab2}
        \end{subfigure}
        \begin{subfigure}[b]{0.15\textwidth}
                \includegraphics[width=\textwidth]{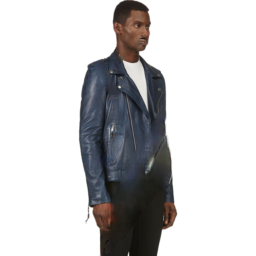}
                \includegraphics[width=\textwidth]{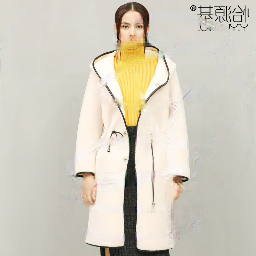}
                \includegraphics[width=\textwidth]{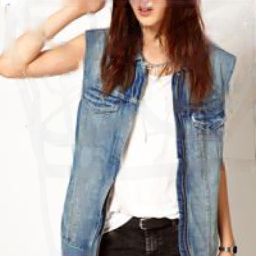}
                \includegraphics[width=\textwidth]{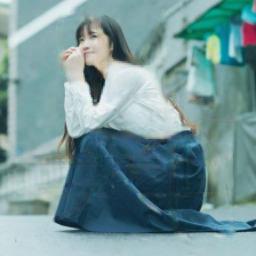}
                \caption{Gated}
                \label{fig:lab3}
        \end{subfigure}
        \begin{subfigure}[b]{0.15\textwidth}
                \includegraphics[width=\textwidth]{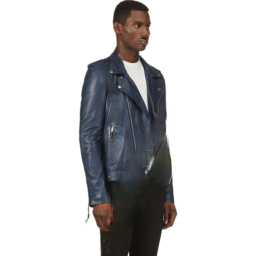}
                \includegraphics[width=\textwidth]{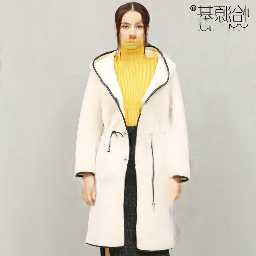}
                \includegraphics[width=\textwidth]{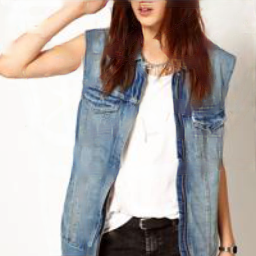}
                \includegraphics[width=\textwidth]{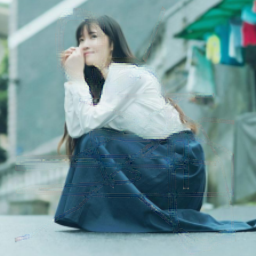}
                \caption{Ours}
                \label{fig:lab4}
        \end{subfigure}
        \begin{subfigure}[b]{0.15\textwidth}
                \includegraphics[width=\textwidth]{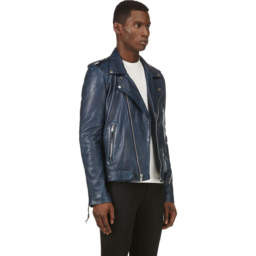}
                \includegraphics[width=\textwidth]{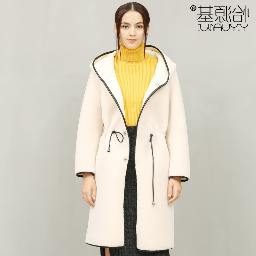}
                \includegraphics[width=\textwidth]{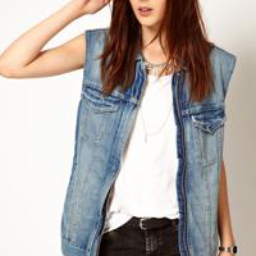}
                \includegraphics[width=\textwidth]{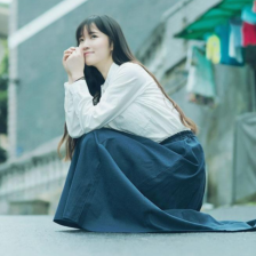}
                \caption{GT}
                \label{fig:lab5}
        \end{subfigure}
        
        \caption{Comparison the results of our proposed model and the state-of-the-art methods.}\label{fig:figure4} 
\end{figure}

Next, we compare the visual compatibility of the results of our model with the other state-of-the-art methods \cite{patchmatch,partial,gated}. Figure \ref{fig:figure4} shows the results on four well-known fashion datasets \cite{fashiongen,fashionai,deepfashion,deepfashion2}. We apply the same settings with ours to the training of all methods, and do not perform any post-processing for the outputs. The results demonstrate that all inpainting strategies can produce visually plausible and semantically coherent clothing images, and the visual outputs of these inpainting strategies can be utilized in order to increase the effectiveness of fashion image understanding solutions (\textit{e.g.} removing the disrupted regions and inpainting them). However, DPConv accomplish it by employing a shallower network architecture, and taking advantage of the efficient mask update step of dilated partial convolutions. Furthermore, to compare the strategies, we can say that (1) DPConv and PConv seem to produce very similar outputs, but when looking into the details, DPConv preserves the visual coherence better (\textit{e.g.} In Figure \ref{fig:figure4},  the collar of the coat in row 2 \& the head of woman in row 4). (2) PM cannot produce smooth outputs in contrast to the other methods. (3) GConv cannot fill the holes by preserving the semantic coherence, and residue of the input mask can be still seen in GConv outputs.

\subsubsection{Ablation Study:} We conduct additional experiments to observe the effect of using dilated convolutions at certain stages of the inpainting networks. The first one is called \textit{DPConv\textsuperscript{\textdagger}} whose layers before dilated partial convolutional block also have dilation of 2 in their kernels. The latter, namely \textit{DGConv}, is the same architecture with our model, but employs gated convolutions, instead of partial convolutions. Table \ref{tab:ablation} demonstrates the quantitative evaluation of these models and ours. The observations are concluded as follows: (1) Using dilation in every stage of the network without applying multi-scale context aggregation strategy has negative effect on the performance of our model. (2) Due to the computational burden, applying multi-scale context aggregation to each stage is not feasible for this task. (3) Gated convolutions with dilation on the kernels leading to the lower-level feature maps shows a similar impact on the qualitative results, which DPConv does it on PConv. (4) However, DPConv still mostly outperforms DGConv on different mask ratios on four well-known fashion datasets.

\begin{table}[!t]
\caption{The evaluation of the usage of dilation in certain stages of the networks for different inpainting strategies.}
\resizebox{\columnwidth}{!}{
\centering
\begin{tabular}{l|l|l|l|l|l|l|l|l|l|l|l|l}
%\toprule
                \textbf{Mask Ratios} & \multicolumn{4}{c|}{\textbf{[0.0:0.2]}}                                                                       & \multicolumn{4}{c|}{\textbf{[0.2:0.4]}}                                                                        & \multicolumn{4}{c}{\textbf{[0.4:]}}                                                                \\ \hline \hline
               \textbf{Datasets} & \multicolumn{1}{c|}{FG} & \multicolumn{1}{c|}{FAI} & \multicolumn{1}{c|}{DF} & \multicolumn{1}{c|}{DF2} & \multicolumn{1}{c|}{FG} & \multicolumn{1}{c|}{FAI} & \multicolumn{1}{c|}{DF} & \multicolumn{1}{c|}{DF2} & \multicolumn{1}{c|}{FG} & \multicolumn{1}{c|}{FAI} & \multicolumn{1}{c|}{DF} & \multicolumn{1}{c}{DF2} \\ 
%\midrule 
\hline \hline
$\ell_{1}$ (DGConv) \textit{\%}   &           0.72         &              0.92           &           0.90            &       \textbf{0.92}                  &          1.17              &         1.43                &        1.21             &         \textbf{1.33}                 &            2.68            &            4.21            &            3.41       &          3.47              \\
$\ell_{1}$ (DPConv\textsuperscript{\textdagger}) \textit{\%}   &         0.62           &       1.02                  &        0.94               &              1.20           &          1.21              &      1.72                   &          1.86            &         2.04                 &        3.18                &        5.16                &        5.66           &           5.39             \\
$\ell_{1}$ (\textbf{DPConv}) \textit{\%}  &        \t
0.70                &         \textbf{0.91}                &        \textbf{0.87}                &         \textbf{0.92}                  &            \textbf{1.14}            &           \textbf{1.39}              &   \textbf{1.20}                    &          \textbf{1.33}                &         \textbf{2.61}                &        \textbf{4.03}                 &       \textbf{3.36}                &      \textbf{3.38}                   \\ 
% \midrule 
\hline \hline
PSNR (DGConv)    &        16.11                &        \textbf{12.31}                 &         12.54               &         \textbf{12.91}                 &         15.09               &         11.76                &            \textbf{11.92}            &          12.10                &           14.12             &          11.56               &            11.70           &          11.92              \\
PSNR (DPConv\textsuperscript{\textdagger})    &       15.96                 &          12.14               &         12.32               &          12.61                &           14.75             &          11.27               &           11.39             &            11.61              &          13.40              &                10.97         &          11.04             &        11.19                \\
PSNR (\textbf{DPConv})   &          \textbf{16.23}              &                   12.27      &          \textbf{12.56}              &       12.89              &  \textbf{15.15}                       &            \textbf{11.81}          &            11.88             &      \textbf{12.12}                    &                 \textbf{14.16}       &       \textbf{11.63}                  &      \textbf{11.84}                  &    \textbf{12.28}                     \\ 
%\midrule 
\hline \hline
SSIM (DGConv)    &       \textbf{0.858}                 &              0.779           &        0.788              &                  0.802        &         0.818               &        0.736                 &        \textbf{0.754}               &          0.774                &          0.750              &         \textbf{0.688}               &        0.712                 &          0.704               \\
SSIM (DPConv\textsuperscript{\textdagger})    &     0.869                   &         0.773                &          0.762            &          0.788                &           0.791             &            0.717             &          0.728             &              0.735            &           0.719             &                 0.626       &            0.664             &       0.679                  \\
SSIM (\textbf{DPConv})   &         0.855        &                \textbf{0.782}         &            \textbf{0.794}            &           \textbf{0.811}              &     \textbf{0.824}               &               \textbf{0.739}          &       0.750                 &        \textbf{0.777}                  &    \textbf{0.766}                    &       0.683                  &      \textbf{0.717}                  &        \textbf{0.719}                 \\ 
\hline\hline
FID (DGConv)  &           3.38             &           5.49              &          4.90              &           5.89               &        \textbf{5.06}                &            10.36             &            9.53            &            10.46              &         22.87               &        25.71                 &        24.02                &         25.88                \\
FID (DPConv\textsuperscript{\textdagger})  &       3.96                 &        6.01                 &        6.73                &         5.82                 &          6.16             &        11.70                 &       11.99                 &            11.38              &         24.02               &         26.89                &        24.54                &          26.24               \\
FID (\textbf{DPConv}) &        \textbf{3.18}                &          \textbf{5.24}              &         \textbf{4.66}               &         \textbf{5.48}            &             5.20           &          \textbf{10.08}              &                \textbf{9.26}        &          \textbf{10.21}               &          \textbf{22.82}              &          \textbf{25.55}               &         \textbf{23.94}                &         \textbf{25.62}                \\ 
%\bottomrule 
\hline \hline
\end{tabular}
}
\label{tab:ablation}
\end{table}

\captionsetup[subfigure]{labelfont=bf, labelformat=parens}
\begin{figure}[!b]
        \centering
        \begin{subfigure}[b]{0.49\textwidth}
                \includegraphics[width=\textwidth]{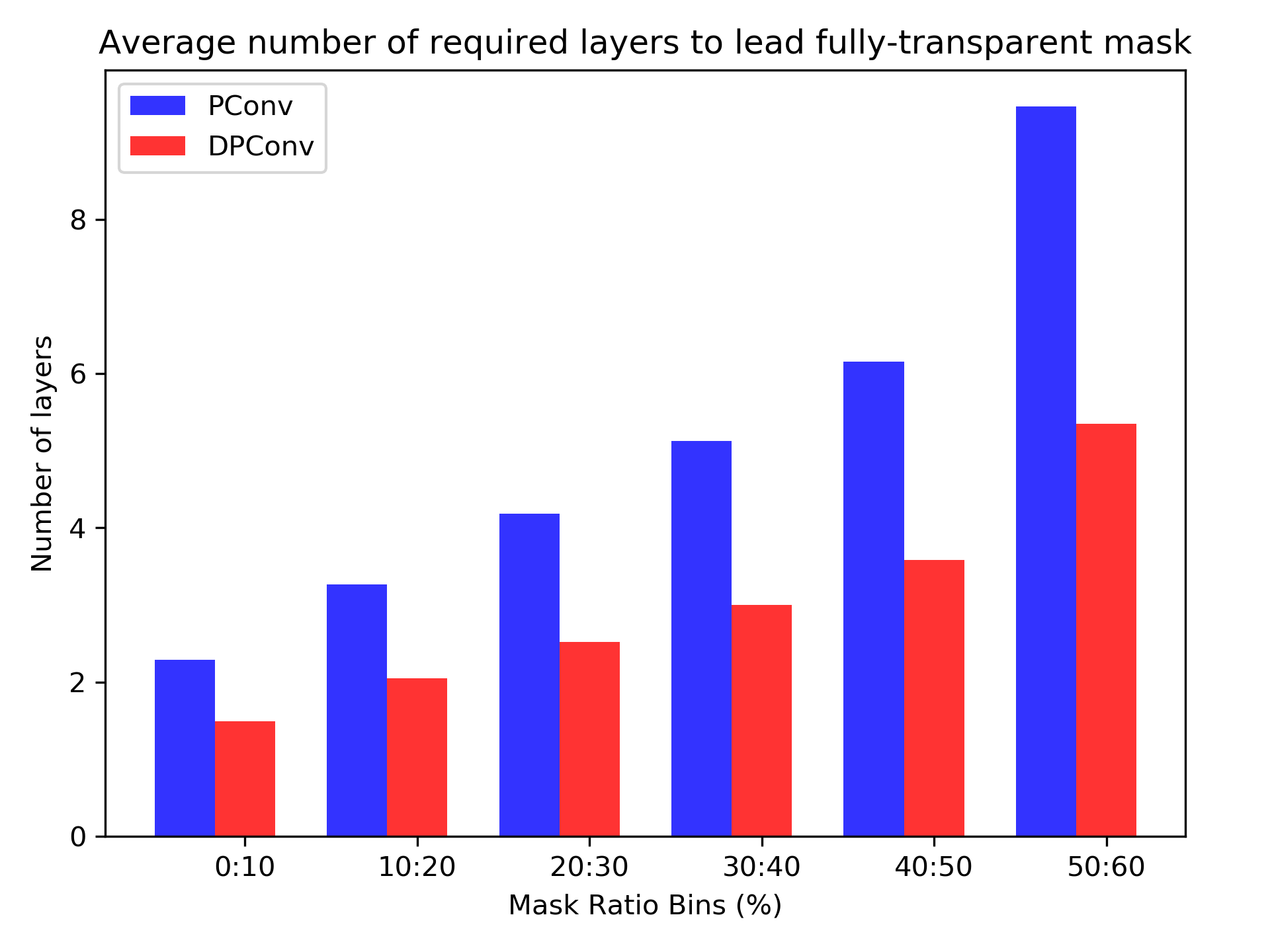}
                \caption{}
                \label{fig:bin_show}
        \end{subfigure}
        \begin{subfigure}[b]{0.49\textwidth}
                \includegraphics[width=\textwidth]{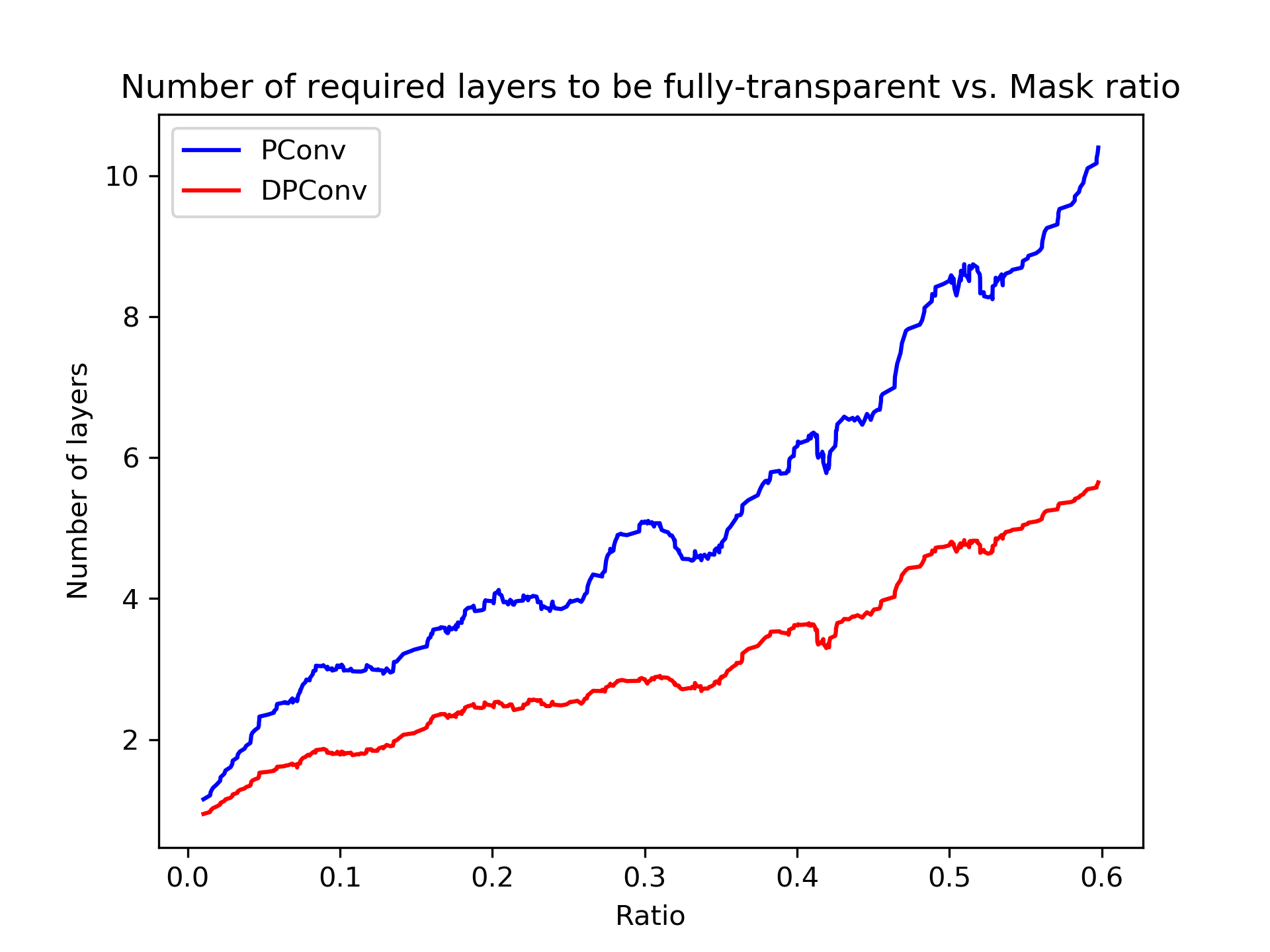}
                \caption{}
                \label{fig:all_show}
        \end{subfigure} 
        
        \caption{Experimental results of analyzing the behavior of partial convolutions and dilated partial convolutions given random masks.}\label{fig:figure-analyze} 
\end{figure}

\subsection{The Effect of Dilation in Partial Convolutions}
Dilated partial convolutions complete the masks to become fully-transparent (\textit{i.e.} masks without any zeros) throughout the network with less number of consecutive layers, when compared to partial convolutions. To empirically prove this, we designed an experiment to analyze the behaviour of both layers given different input masks. In this experiment, we used 12.000 random masks with {\raise.17ex\hbox{$\scriptstyle\sim$}}2.000 masks for each 10\% range of mask ratios from 0\% up to 60\%. For each range of mask ratios, the average number of layers required to obtain a fully-transparent mask is calculated, and the maximum number of layers that can be stacked is limited to 20. As illustrated in Figure \ref{fig:bin_show}, dilated partial convolutions can reach fully-transparent masks 2.1 layers earlier on average (\textit{i.e.} {\raise.17ex\hbox{$\scriptstyle\sim$}}15\% less layers). Figure \ref{fig:all_show} emphasizes the exponential growth of the required number of layers with respect to all masks in the case of using partial convolutions or dilated partial convolutions. 
% Note that we applied Savitzky-Golay filter \cite{savgol} to the lists of number of required layers for both layer types, in order to be able to fit a smoother graph with 12.000 mask instances. 
The result of this experiment shows that the dilated version of partial convolutions has a practical impact of \textbf{leading fully-transparent masks in higher resolution without requiring to go deeper throughout the network}, and thus it leads to a more efficient mask update step and faster learning process. Note that PConv decreases the feature map to $2x2$ in its decoder part while DPConv starts to upsampling at the feature maps sized $32x32$.

\captionsetup[subfigure]{labelfont=bf, labelformat=empty}
\begin{figure}[!b]
        \centering
        \begin{subfigure}[b]{0.16\textwidth}
                \includegraphics[width=\textwidth]{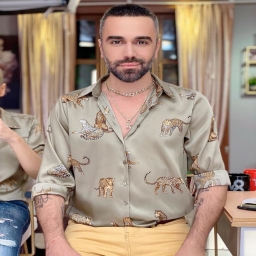}
                \includegraphics[width=\textwidth]{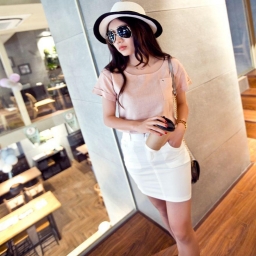}
               \includegraphics[width=\textwidth]{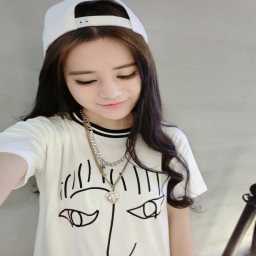}
               \caption{GT}
        \end{subfigure}       
        \begin{subfigure}[b]{0.16\textwidth}
                \includegraphics[width=\textwidth]{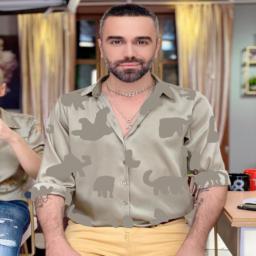}
                \includegraphics[width=\textwidth]{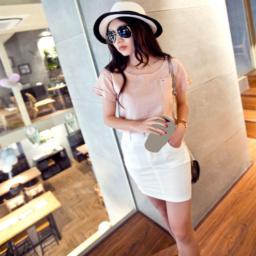}
                \includegraphics[width=\textwidth]{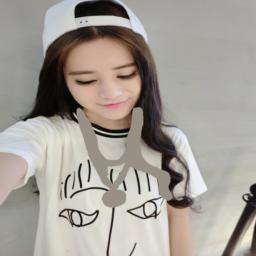}
                \caption{IN}
        \end{subfigure}
        \begin{subfigure}[b]{0.16\textwidth}
                \includegraphics[width=\textwidth]{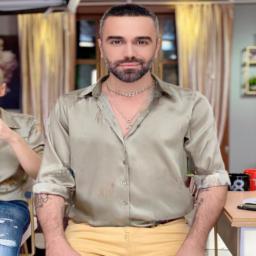}
                \includegraphics[width=\textwidth]{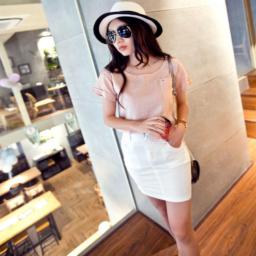}
               \includegraphics[width=\textwidth]{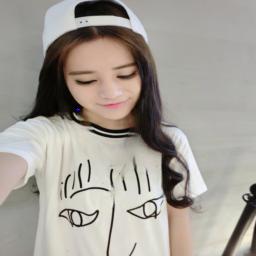}
               \caption{OUT}
        \end{subfigure} 
        \begin{subfigure}[b]{0.16\textwidth}
                \includegraphics[width=\textwidth]{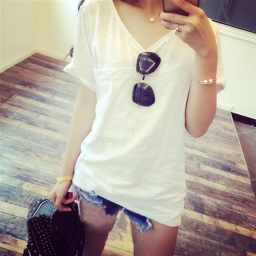}
                \includegraphics[width=\textwidth]{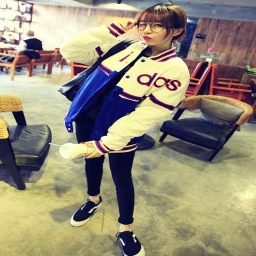}
               \includegraphics[width=\textwidth]{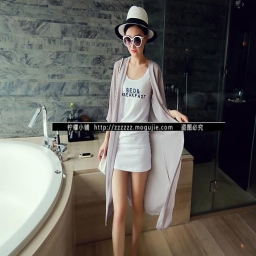}
               \caption{GT}
        \end{subfigure} 
        \begin{subfigure}[b]{0.16\textwidth}
                \includegraphics[width=\textwidth]{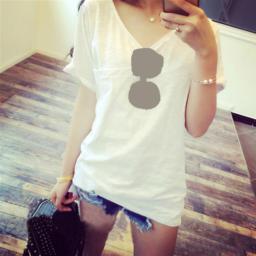}  \includegraphics[width=\textwidth]{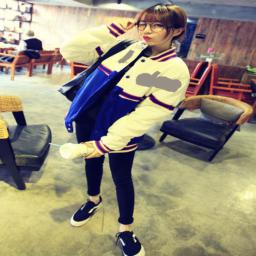}
                \includegraphics[width=\textwidth]{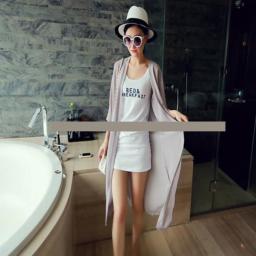}
                \caption{IN}
        \end{subfigure}
        \begin{subfigure}[b]{0.16\textwidth}
                \includegraphics[width=\textwidth]{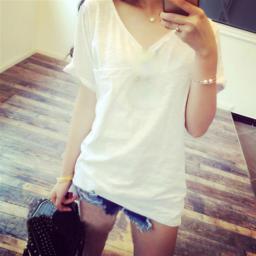}
                \includegraphics[width=\textwidth]{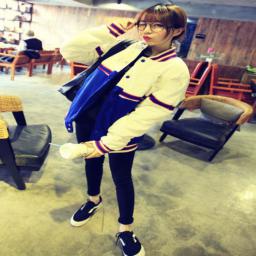}
                \includegraphics[width=\textwidth]{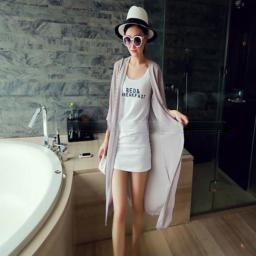}
                \caption{OUT}
        \end{subfigure}
        \caption{Example inpainting results of the images containing some overlapping or disruptive parts on clothing items.}\label{fig:figure-ablation} 
\end{figure}

\subsection{Practical Usage in Fashion Domain}

Image inpainting can improve the performance of fashion image understanding solutions when applied as pre-process or post-process. To give a brief idea about how it can be useful for such systems, we picked a number of images that contain different items overlapping the clothing items or that have some disruptive parts (\textit{e.g.} banner, logo). We have manually created the masks that remove these items/parts from the images. Figure \ref{fig:figure-ablation} demonstrates the inpainting results of picked images where DPConv trained on DeepFashion2 dataset is employed for testing. At this point, we showed that clothing image inpainting can be useful for fashion editing (\textit{e.g.} removing accessories like eye-glasses and necklace, logos, banners or non-clothing items, even changing the design of clothing), and it makes the main-stream fashion image understanding solutions work better.

\captionsetup[subfigure]{labelfont=bf, labelformat=empty}
\begin{figure}[!t]
        \centering
        \begin{subfigure}[b]{0.16\textwidth}
                \includegraphics[width=\textwidth]{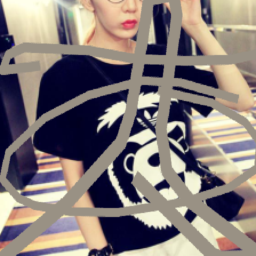}
                \includegraphics[width=\textwidth]{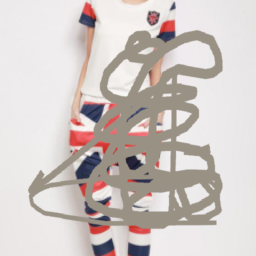}
                \includegraphics[width=\textwidth]{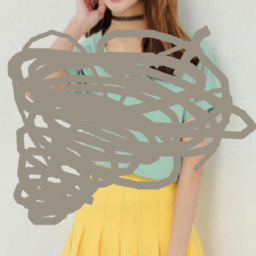}
                \includegraphics[width=\textwidth]{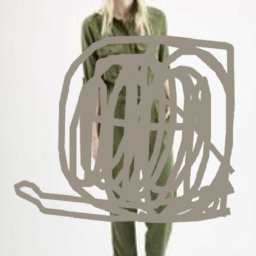}
                \caption{IN}
        \end{subfigure}
        \begin{subfigure}[b]{0.16\textwidth}
                \includegraphics[width=\textwidth]{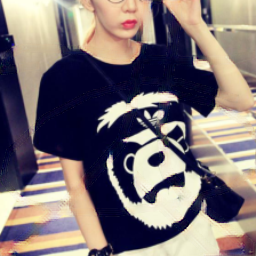}
                \includegraphics[width=\textwidth]{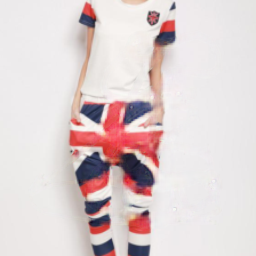}
                \includegraphics[width=\textwidth]{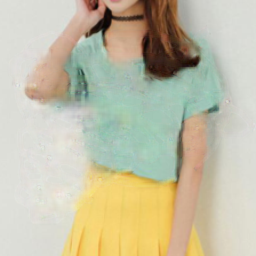}
                \includegraphics[width=\textwidth]{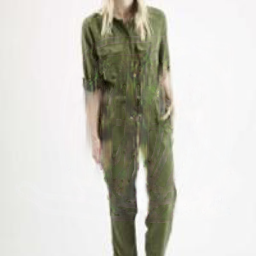}
                \caption{OUT}
        \end{subfigure}
        \begin{subfigure}[b]{0.16\textwidth}
                \includegraphics[width=\textwidth]{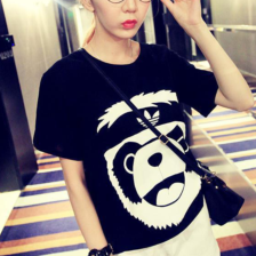}
                \includegraphics[width=\textwidth]{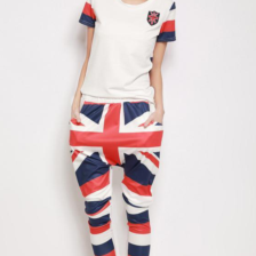}
                \includegraphics[width=\textwidth]{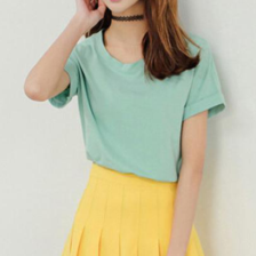}
                \includegraphics[width=\textwidth]{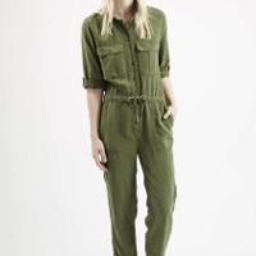}
                \caption{GT}
        \end{subfigure}
        \begin{subfigure}[b]{0.16\textwidth}
                \includegraphics[width=\textwidth]{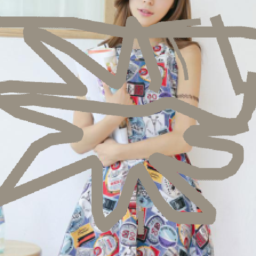}
                \includegraphics[width=\textwidth]{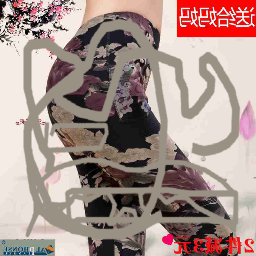}
                \includegraphics[width=\textwidth]{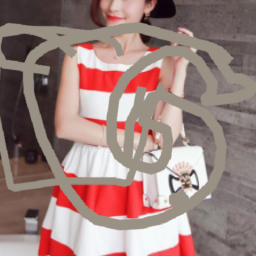}
                \includegraphics[width=\textwidth]{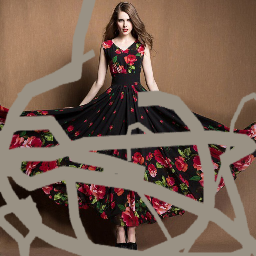}
                 \caption{IN}
        \end{subfigure}
        \begin{subfigure}[b]{0.16\textwidth}
                \includegraphics[width=\textwidth]{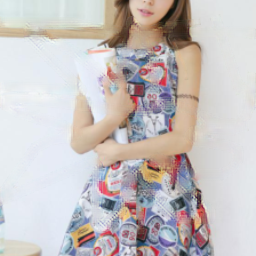}
                 \includegraphics[width=\textwidth]{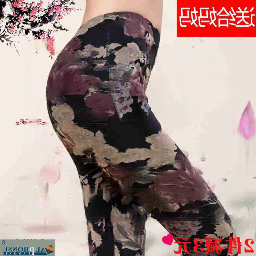}
                 \includegraphics[width=\textwidth]{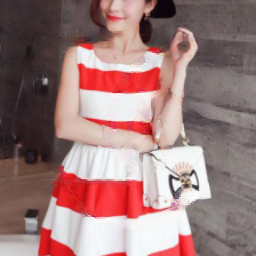}
                \includegraphics[width=\textwidth]{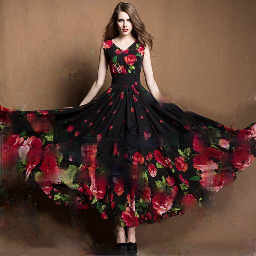}
                  \caption{OUT}
        \end{subfigure}
        \begin{subfigure}[b]{0.16\textwidth}
                \includegraphics[width=\textwidth]{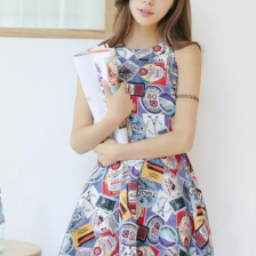}
                \includegraphics[width=\textwidth]{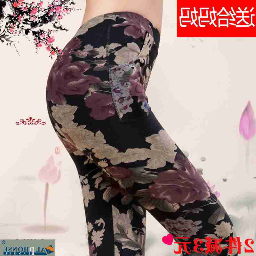}
                \includegraphics[width=\textwidth]{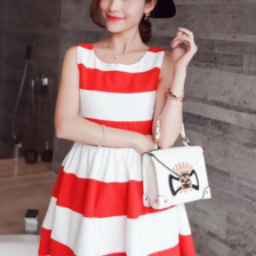}
                \includegraphics[width=\textwidth]{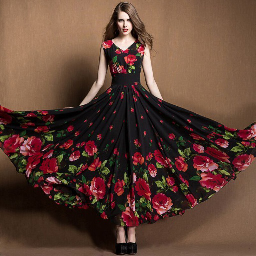}
                 \caption{GT}
        \end{subfigure}
        
        \caption{More inpainting results of DPConv on four different datasets. Zoom in for better view.}
        \label{fig:figure7} 
\end{figure} 

\section{Conclusion}
In this study, we present an extensive benchmark for clothing image inpainting, which may be practical for industrial applications in fashion domain. 
%In image inpainting literature, the main research focus is on facial image understanding and natural scene understanding, and this study mainly aims to re-direct the focus to fashion image understanding to reveal its potential. 
Qualitative and quantitative comparisons demonstrate that proposed method improves image inpainting performance when compared to the previous state-of-the-art methods, and produce visually plausible and semantically coherent results for clothing images. Overall performances of inpainting strategies proves that AI-based fashion image understanding solutions can employ inpainting to their pipeline in order to improve the general performance. 

\clearpage


\bibliographystyle{splncs04}
\bibliography{eccv2020submission.bib}
\end{document}